\documentclass[letterpaper, 10pt, journal, twoside]{IEEEtran}
\IEEEoverridecommandlockouts
\usepackage{graphicx}
\usepackage{amssymb}
\usepackage{amsmath}
\usepackage{multirow}
\usepackage{bbm}
\usepackage[size=footnotesize]{caption}
\usepackage{subcaption}

\usepackage{xcolor}

\usepackage{pifont}
\newcommand{\cmark}{\ding{51}}%
\newcommand{\xmark}{\ding{55}}%

\title{CLONeR: \textbf{C}amera-\textbf{L}idar Fusion for \textbf{O}ccupancy Grid-aided \textbf{Ne}ural \textbf{R}epresentations}

\author{Alexandra Carlson$^{*,1}$, Manikandasriram S. Ramanagopal$^{*,2}$, \\ Nathan Tseng$^1$, Matthew Johnson-Roberson$^3$, Ram Vasudevan$^2$, and
    Katherine A. Skinner$^2$
    \thanks{$^*$denotes equal contribution}%
    \thanks{Manuscript received: January 04, 2023; Accepted March 08, 2023. This paper was recommended for publication by Editor Cesar Cadena upon evaluation of the Associate Editor and Reviewers' comments.
    This work was supported by a grant from Ford Motor Company via the Ford-UM Alliance under award N028603.}% Use only for final RAL version
    \thanks{$^1$A. Carlson and N. Tseng are with Ford Motor Company, but completed this work during graduate studies at the Department of Robotics, University of Michigan, Ann Arbor, MI, USA {\tt\footnotesize askc,tsnathan@umich.edu}.}%
    \thanks{$^2$M. Ramanagopal, R. Vasudevan, and K. Skinner are with the Department of Robotics, University of Michigan, Ann Arbor, MI, USA {\tt\footnotesize srmani,ramv,kskin@umich.edu}.}%
    \thanks{$^3$M. Johnson-Roberson is with the Robotics Institute, Carnegie Mellon University, Pittsburgh, PA, USA {\tt\footnotesize mkj@andrew.cmu.edu}.}%
    \thanks{Website: https://fcav.engin.umich.edu/cloner}%
    \thanks{Digital Object Identifier (DOI): 10.1109/LRA.2023.3262139.}%
}
\begin{document}

\maketitle

% \thispagestyle{empty}
% \pagestyle{empty}
% Comment or remove these lines for final RAL version.

%===============================================================================

\begin{abstract}
    Recent advances in neural radiance fields (NeRFs) achieve state-of-the-art novel view synthesis and facilitate dense estimation of scene properties. However, NeRFs often fail for outdoor, unbounded scenes that are captured under very sparse views with the scene content concentrated far away from the camera, as is typical for field robotics applications. In particular, NeRF-style algorithms perform poorly: (1) when there are insufficient views with little pose diversity, (2) when scenes contain saturation and shadows, and (3) when finely sampling large unbounded scenes with fine structures becomes computationally intensive.
   This paper proposes \textit{CLONeR}, which significantly improves upon NeRF by allowing it to model large unbounded outdoor driving scenes that are observed from sparse input sensor views. This is achieved by decoupling occupancy and color learning within the NeRF framework into separate Multi-Layer Perceptrons (MLPs) trained using LiDAR and camera data, respectively. In addition, this paper proposes a novel method to build differentiable 3D Occupancy Grid Maps (OGM) alongside the NeRF model, and leverage this occupancy grid for improved sampling of points along a ray for volumetric rendering in metric space. 
    Through extensive quantitative and qualitative experiments on scenes from the KITTI dataset, this paper demonstrates that the proposed method outperforms state-of-the-art NeRF models on both novel view synthesis and dense depth prediction tasks when trained on sparse input data.
\end{abstract}

% Keywords appear just beneath the abstract. Use only for final RAL version. 
% Check list here for better keywords: https://www.ieee-ras.org/publications/ra-l/keywords
\begin{IEEEkeywords}
Deep Learning for Visual Perception, Sensor Fusion, Computer Vision for Transportation
\end{IEEEkeywords}
\section{Introduction}
\label{sec:intro}

% APPLICATION/MOTIVATION
% Drop letter for first word of the Introduction
% Use only for final RAL versionFormatting
\IEEEPARstart{T}{he} estimation of dense scene properties from sparse and low diversity sensor data is a critical task in robot perception. In particular, for autonomous vehicle perception, estimating dense and accurate depth is essential for scene understanding used for downstream planning and navigation tasks. Neural radiance fields (NeRFs) are an attractive solution for producing joint dense depth and color maps of a scene~\cite{mildenhall2020nerf} . NeRFs use implicit functions to encode volumetric density and color observations from multiple images with known poses. Recent advances in NeRFs have demonstrated state-of-the-art performance for novel view synthesis and 3D modeling.

% KEY GAP/LIMITATION IN CURRENT METHODS
Still, outdoor scenes captured in autonomous vehicle datasets remain challenging for NeRFs. These scenes are typically captured from sparse viewpoints with limited view diversity, i.e., a camera mounted on a vehicle moving forward. They tend to be large and unbounded, with 
much of the scene content concentrated far away from the camera. Lastly, images captured outdoors feature varying exposure and brightness levels depending upon the camera settings and weather conditions. These challenges can lead to learning inaccurate scene geometry, blurry scene renderings and poorly modeled thin structures, especially at long distances. 

\begin{figure}[t]
    \centering
    \includegraphics[width=\linewidth]{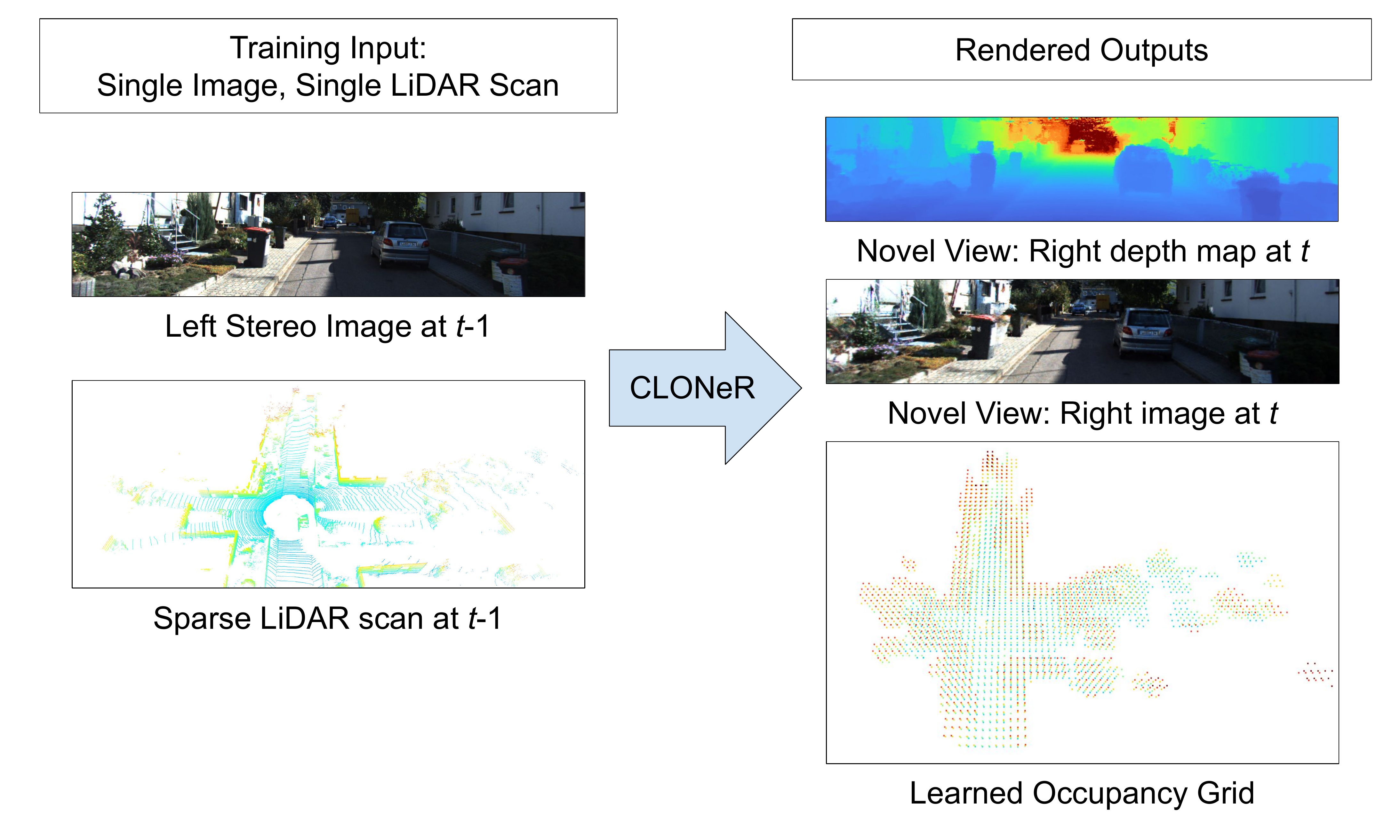}
    \caption{An illustration of the sparse inputs and dense outputs from CLONeR.  From a single image and single LiDAR scan as input (shown on the left, under `Training Inputs'), CLONeR (middle, represented by the blue arrow) can generate accurate dense depth, novel views, and an Occupancy grid map (shown on the right, under `Rendered Outputs').} %
    \label{fig:introfig}
\end{figure} 

% CONTRIBUTIONS
This paper presents \textbf{\textit{CLONeR}}, \textbf{C}amera-\textbf{L}idar fusion for \textbf{O}ccupancy grid-aided \textbf{Ne}ural \textbf{R}epresentation, a novel NeRF framework that addresses key challenges to applying NeRFs to outdoor driving scenes. The proposed method can accurately \textit{clone} a 3D scene from a sparse set of image views and LiDAR scans. Impressively, CLONeR can render novel views and dense depth maps with fine detail from a single image and LiDAR scan (see Fig.~\ref{fig:introfig}). CLONeR accurately models the operation of both LiDAR and RGB camera sensors within a NeRF framework to learn a dense 3D representation of a scene. In addition, CLONeR simultaneously builds a differentiable 3D Occupancy Grid Map (OGM) using LiDAR data, which it leverages to efficiently sample large outdoor scenes in metric space. 
This learned occupancy grid map is capable of maintaining known versus unknown regions in a probabilistic manner, and replaces the standard coarse MLP sampling scheme used in NeRFs. Critically, CLONeR decouples occupancy and color learning within the NeRF by utilizing two separate MLPs. One of these MLPs learns occupancy from LiDAR ray data, and the other MLP learns the RGB model of the scene from camera rays. By doing this, sensor artifacts and illumination effects in the RGB images do not negatively influence the learned scene depth.

In summary, the key contributions of the proposed work, CLONeR, are:
\begin{itemize}
    \item A decoupled NeRF model where LiDAR is used to supervise geometry learning while camera images are used to supervise color learning. This allows one to perform novel view synthesis using as few as a single image when LiDAR scans are available.
    \item A differentiable 3D occupancy grid learned directly on the GPU alongside the decoupled NeRF model. This 3D occupancy grid explicitly maintains information about known and unknown regions, and outperforms standard coarse NeRF based importance sampling methods.
\end{itemize}
\noindent We demonstrate that our method outperforms state-of-the-art NeRF models on both novel view synthesis and depth prediction tasks through extensive quantitative and qualitative experiments on several scenes from the KITTI dataset. 

The remainder of this paper is organized as follows: We review related work in Section~\ref{sec:background}, and give a detailed description of the proposed method, CLONeR, in Section~\ref{sec:methods}. We present extensive quantitative and qualitative results in Section~\ref{sec:results}. Section~\ref{sec:conclusion} provides the conclusion and future work.

\section{Related Works}
\label{sec:background}

The impressive performance achieved by the original NeRF~\cite{mildenhall2020nerf} has inspired many subsequent works to extend it to various applications including generative modeling \cite{chan2021pi} and scene relighting \cite{zhang2021nerfactor}. 
In this work, we develop a novel NeRF framework that leverages camera and LiDAR to enable depth-guided learning for unbounded, outdoor scenes observed from sparse/limited viewpoints.

\subsection{Depth-guided Learning in Neural Radiance Fields}

In prior work, NeRF-based models have been used along with depth supervision from structure-from-motion techniques \cite{deng2022depth}, RGB-D cameras \cite{sucar2021imap, Zhu2022CVPR,neff2021donerf} and continuous-wave Time-of-Flight sensors \cite{attal2021torf}. 
However, RGB-D cameras do not operate well outdoors in the presence of sunlight, and SfM pipelines operate poorly in low-texture environments, making this depth supervision very difficult to obtain for outdoor scenes that would be encountered in autonomous driving applications.

\subsection{Neural radiance fields for outdoor scenes}
A significant number of NeRF models have tried to tackle more complex static 3D objects, as well as larger indoor scenes~\cite{sucar2021imap,piala2021terminerf,wang2021neus,yariv2021volume,niemeyer2022regnerf}. In particular, in the surface reconstruction problem space, both NeuS~\cite{wang2021neus} and VolSDF~\cite{yariv2021volume} both separate the color and opacity learning. We extend their intuitions to volume rendering in this work, and demonstrate that decoupling is key in removing several failure modes when training NeRFs on outdoor scenes.
However, outdoor scenes still present a unique challenge for NeRFs. Many outdoor scenes have a relatively large and unbounded. Much of the scene content is concentrated far away from the camera(s) with a significant amount of empty space. The scenes are large enough that sampling the 3D scene volume to learn accurate scene structures becomes computationally challenging~\cite{neff2021donerf}. As a result, NeRFs struggle to learn even coarse scene occupancy. This results in blurry scene renderings and poorly modeled fine/thin structures, especially at long distances~\cite{zhang2020nerf++, barron2022mip}. 

There have been many successful methods, including BlockNerF~\cite{tancik2022block} and CityNeRF~\cite{xiangli2021citynerf}, that focus on large scale scenes that cover a large spatial extent, such as a city block or entire city. These methods leverage multiple NeRF models to learn to model the 3D space~\cite{tancik2022block} and they require a significant number of images to train. In our proposed method, we instead focus on scenes with a long depth range from a single camera view, which is specific to the type of data we would encounter in autonomous vehicle datasets. Additionally, we consider a single scene from a limited viewpoint and develop our novel contributions in the OGM and decoupled models for handling LiDAR for these cases. This differs greatly in the configuration of the scenes in works such as BlockNeRF, which focuses on combining smaller NeRF models to handle the large spatial extent of their chosen scenes. There has also been very promising work in modeling 360$^{\circ}$ unbounded scenes~\cite{barron2021mip360, zhang2020nerf++,martin2021nerf,rematas2022urban}, but these methods require dense multiview datasets which are difficult to capture for autonomous vehicle datasets.

Very few NeRF models have tried to use LiDAR information to learn scene occupancy in outdoor scenes. Of particular relevance to our work is Urban Radiance Fields \cite{rematas2022urban} and Neural Point Light Fields~\cite{ost2022neural}. The method proposed in Urban Radiance fields utilizes LiDAR data along with images from panoramic cameras captured in outdoor scenes. Their data collection trajectory specifically ensures a large number of views are available with diversity in the camera poses in addition to their LiDAR data. In comparison, the proposed method is designed to operate on very sparse camera views and LiDAR data~\cite{rematas2022urban}. 

In Neural Point Light Fields, they represent outdoor scenes with a light field learned on top of a sparse point cloud. They aggregate point features along each ray to get a final RGB prediction. However, this method learns a feature vector (via CNN encoding) for each point in the point cloud, and thus requires long driving trajectories/videos to learn a sufficient scene representation~\cite{ost2022neural}. In contrast, our proposed method can produce accurate scene models from as few as a single image paired with a single point cloud. 
In summary, neither ~\cite{rematas2022urban,ost2022neural} have demonstrated the ability to learn dense depth using just LiDAR returns and sparse views.

\subsection{Sampling in Neural Radiance Fields}

A key challenge in the original NeRF formulation is the need to finely sample along a ray to approximate the line integral. 
Mip-NeRF \cite{barron2021mip} addressed the sampling and aliasing challenges in the original NeRF formulation and Mip-NeRF 360 \cite{barron2021mip360} extended it to address unbounded scenes with focus on a central object. 
TermiNeRF \cite{piala2021terminerf} and DO-NeRF \cite{neff2021donerf} replace the coarse MLP of the original NeRF model with a separate sampling MLP that estimates where the fine MLP network should be applied in the 3D world volume. However, in order to train the aforementioned networks, the input data still needs to cover a large and diverse set of views. 

Recent developments have popularized the use of occupancy grids for sampling during test-time in NeRFs~\cite{yu2021plenoctrees,sun2022improved}. 
Most similar to ours, in \cite{mueller2022instant}, the authors propose to maintain a cascaded set of occupancy grids in order to skip ray marching on empty space. They present heuristic methods to update the occupancy grids as training of the NeRF model progresses. In contrast, 
we propose to jointly perform occupancy grid mapping while training the NeRF model. We further distinguish our approach by leveraging LiDAR measurements to learn and construct these occupancy grids in a probabilistic manner. 

\section{Methods}
\label{sec:methods}

\begin{figure*}
    \centering
    \includegraphics[width=\linewidth]{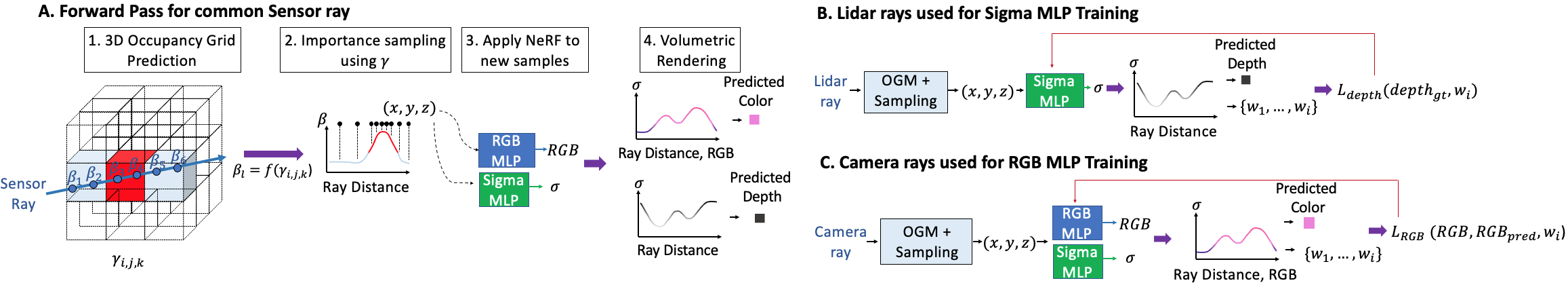}
    \caption{Overview of proposed CLONeR framework. Panel A shows the forward pass of the framework. In Panel B, LiDAR data is used as the only supervision for the Sigma MLP during training. In Panel C, camera data is the only supervision used to train the Color MLP.}
    \label{fig:methodoverview}
\end{figure*}

Figure~\ref{fig:methodoverview} provides an overview of the proposed method. Its two key components are (i) the differentiable occupancy grid map (OGM), which is used to learn the coarse occupancy of the 3D metric space of the scene, and (ii) the decoupled NeRF MLPs, which independently learn a fine color and depth model of the scene. We describe each model component and the training process of the framework in the following subsections.

\subsection{Overview of Neural Radiance Fields}
\label{subsec:nerf_primer}

A Neural Radiance Field (NeRF)~\cite{mildenhall2020nerf} models the scene using a 5D vector-valued function using an MLP network. This MLP takes a 3D location $\vec{x} = (x,y,z)$ and a viewing direction $\vec{d} = (x_d, y_d, z_d)$ as inputs and outputs the color $\vec{c} = (r, g, b)$ emitted by that scene point and a scalar volume density $\sigma$. 

An image pixel $(u, v)$ is mapped to a camera ray $\vec{r}(t) = \vec{o} + t\vec{d}$ using a simple pinhole camera model. The expected color of the camera ray is computed using the quadrature approximation of the volumetric rendering equation between the near and far bounds $t_n$ and $t_f$~\cite{mildenhall2020nerf}:
\begin{equation}
\begin{split}
    \vec{C}(\vec{r}) = \sum_{i=1}^N T_i (1-\exp(-\sigma_i \delta_i)) \vec{c}_i, \quad \\
    \mathrm{where} \quad T_i = \exp \big( -\sum_{j=1}^{i-1} \sigma_j \delta_j \big),
    \label{eq4:line_integral_color}
\end{split}
\end{equation}

\noindent and $\delta_i = (t_{i+1} - t_i)$ is the distance between samples along the ray where the MLP is applied. The weights of the MLP are then optimized by minimizing the mean squared error between the rendered color $\hat{\vec{C}}(\vec{r})$ of a pixel and the ground truth color of that pixel.
The termination depth of a ray $\hat{D}(\vec{r})$ is computed as:
\begin{equation}
\begin{split}
    \hat{D}(\vec{r}) = \sum_{i=1}^N T_i (1-\exp(-\sigma_i \delta_i)) t_i.
    \label{eq4:line_integral_depth}
\end{split}
\end{equation}

\subsection{CLONeR World Coordinate System}

Given the set of camera and LiDAR poses from an external source such as a GNSS system, we follow \cite{mildenhall2020nerf} and subtract the average translation and normalize the rotation such that the local $z-$axis for each camera points into the camera. The sensor poses and the point clouds from the LiDAR are further scaled by a common factor such that the region of interest falls completely within $[-1, 1]^3$, as required by most types of positional encodings used in NeRFs \cite{mildenhall2020nerf}. We refer to this scaled, bounded domain as the \emph{world cube}. 

For generic outdoor driving scenes, we determine the region of interest in the following way: for each input camera pose, we use its intrinsic parameters and user-defined near and far values to compute the 8 corners of its view frustum in world coordinates. 
We choose a set of desired camera poses along with their intrinsics and compute corresponding view frustum points. We then compute a common scale factor such that all the sensor poses in addition to all the above points fall within the world cube. We explicitly avoid using the LiDAR point clouds to define the scale factor as LiDAR data could contain regions never seen in any of the images, such as points behind the camera. 

\subsection{Integrating LiDAR Measurements into NeRF}

A 3D scanning LiDAR continuously sweeps the scene using a set of laser beams. 
Let $H_L$ denote the pose of the LiDAR in world coordinates (as a $3\times4$ matrix) and let $\vec{x}$ denote a 3D point measured in the local sensor frame at some time $t$. 
This corresponds to a ray in the world coordinates with ray origin $\vec{o} = H_L [0, 0, 0, 1]^T$ and ray direction $\vec{d} = H_L [\vec{x}^T 1]^T - \vec{o}$.
Recall from Sec.~\ref{subsec:nerf_primer} that one can render a camera ray using the NeRF MLP and obtain a termination depth, expected color, as well as opacity at samples along the ray. Similarly, based upon how we model the LiDAR ray, we can also render these quantities for each LiDAR ray cast into the scene.

\subsection{Decoupled NeRF Model}

We propose a decoupled NeRF model that contains two separate MLP networks, one for learning the scene geometry and one for scene color: $F^g_{\Theta} : \mathrm{R}^3 \rightarrow \mathrm{R}$, which we call the sigma MLP, and $F^c_{\Phi} : \mathrm{R}^5 \rightarrow \mathrm{R}^{n_c}$,which we call the color MLP. Note that $n_c$ is the number of color channels. These decoupled MLP networks are trained simultaneously such that $\sigma_i = F^g_{\Theta} (\vec{r}(t_i))$ and $\vec{c}_i = F^c_{\Phi} (\vec{r}(t_i), \vec{d})$.

These two networks are shown in panels B and C of Fig.~\ref{fig:methodoverview}.
Our sigma MLP takes the positionally encoded 3D positions only as input. It has 1 hidden layer of 64 neurons with the estimated differential opacity $\sigma$ as output.
Our color MLP takes the positionally encoded 3D positions and encoded view directions as input. It has 2 hidden layers of 64 neurons each and outputs colors with $n_c$ channels. 
We use multi-resolution hash tables~\cite{mueller2022instant} for positional encoding and use Spherical Harmonics \cite{yu2021plenoctrees} for encoding view direction.
We use separate hash tables for positional encoding of the sigma and color MLPs. Unlike the original NeRF, there is no interaction between the parameters of the sigma MLP and the color MLP.

When rendering a LiDAR ray, we only evaluate the sigma MLP while computing gradients.
When rendering a camera ray, we evaluate the sigma MLP, without computing gradients, to estimate the differential opacity needed in Eq.~\ref{eq4:line_integral_color} to compute the expected color. In contrast to the original NeRF model, which uses a simple pinhole camera model, we instead use the full pinhole camera model to compute ray directions. This is needed to maintain consistency with camera-LiDAR calibration parameters. We use multiple samples per pixel by generating rays corresponding to fractional pixel coordinates. To obtain ground truth color information for the fractional pixels, we use bilinear interpolation of the input images.

\subsection{Loss Functions for NeRF MLPs}

We use separate loss functions for LiDAR and camera rays: 
\paragraph{LiDAR Line-of-sight (LOS) loss} We use a variant of the line-of-sight loss used in \cite{rematas2022urban}. For a given LiDAR ray, let $z^*$ be the ground truth termination depth along the ray. Let $t_i$ be the samples along the ray and $w_i$ be the associated accumulated opacity at those samples. We define $w_i^* = \mathcal{K}_\epsilon (t_i - z^*)$ where $\mathcal{K}_{\epsilon}$ is a truncated Gaussian distribution with variance equal to $(\epsilon/3)^2$ and $\epsilon$ is a non-zero number that is decayed as training progresses.
We can now define the line-of-sight loss as
\begin{equation}
    \mathcal{L}_{\mathtt{sight}}(\Theta) = \|w_i - w_i^*\|_1
\end{equation}
Ideally, we would like $w_i$ to be non-zero only at a single point and zero everywhere else. 
To encourage this sparse behavior, we use an L1 penalty. 

\paragraph{LiDAR Opacity loss} The volumetric rendering equation as implemented in the original NeRF model only requires that the sum of accumulated opacity does not exceed 1.
However, Eq.~\ref{eq4:line_integral_depth} implicitly assumes that the accumulated opacities sum to 1. 
Therefore, we use the penalty method to force the accumulated opacity for a ray to sum to 1 by using the following opacity loss:
\begin{equation}
    \mathcal{L}_{\mathtt{opacity_l}}(\Theta) = \| 1 - \sum_i w_i \|
\end{equation}
where $\Theta$ is the trainable parameters of the sigma MLP.

\paragraph{Camera Color loss} Following \cite{sucar2021imap}, we use L1 penalty for supervising camera rays.

\begin{equation}
    \mathcal{L}_{\mathtt{color}}(\Phi) = \|\hat{\vec{C}}(\vec{r}) - \vec{C}(\vec{r}) \|_1
\end{equation}

\paragraph{Total Loss} The complete loss function is given by
\begin{equation}
    \mathcal{L} (\Theta, \Phi) = \lambda_1 \mathcal{L}_{\mathtt{sight}}(\Theta) +\mathcal{L}_{\mathtt{opacity_l}}(\Theta) + \\
    \lambda_2 \mathcal{L}_{\mathtt{color}}(\Phi)
\end{equation}
where $\lambda_1 = 1000$ and $\lambda_2 = 1$, and only $\lambda_1$ is decayed over training iterations.
%% OGM OVERVIEW FIGURE
\begin{figure}
    \centering
    \includegraphics[width=\linewidth]{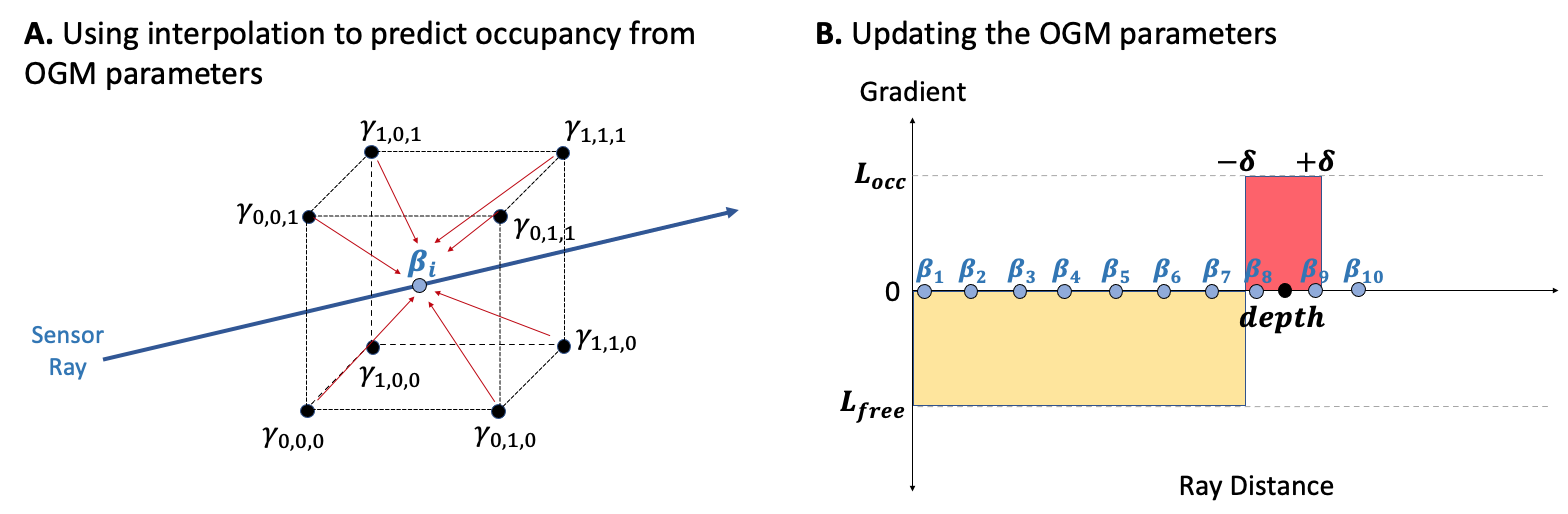}
    \caption{Occupancy Grid Mapping as Stochastic Gradient Descent. Panel A describes the interpolation process to predict occupancy values at each ray sample from the neighboring $\gamma$ variables in the OGM.  Panel B describes how the $\gamma$ variables of the OGM are updated using stochastic gradient descent.}
    \label{fig:methods_occ_grid}
\end{figure}

\subsection{Occupancy Grid Mapping}
Occupancy Grid Mapping (OGM) has long been used in robotics to model 3D space in a probabilistic manner~\cite{thrun2002probabilistic}. In this paper, we cast OGM as an optimization problem that can be solved using stochastic gradient descent. 
We then use the learned OGM to efficiently sample along sensor rays, taking the place of a coarse MLP in standard NeRF algorithms. 

Let $\mathcal{O} \in R^{N\times N\times N}$ denote a regular 3D grid partitioning a bounded region of interest into voxels. If $p_k$ refers to the probability that a single grid element is occupied, then the corresponding log-odds is defined as $l_k = \log \frac{p_k}{1-p_k}$.
In a typical OGM algorithm, at each iteration $n$, given a range measurement $z_n$, each grid element is updated as follows:
\begin{equation}
    l_{n, k} = l_{n-1, k} + \mathcal{S}(k, z_n) - l_0
    \label{eq4:logodds_update}
\end{equation}
where $l_0$ is the prior of occupancy represented in log-odds, $\mathcal{S}$ is the inverse sensor model that returns the measurement update for grid element $k$ given $z_n$ and the known pose of the sensor~\cite{thrun2002probabilistic}. 
We propose to treat the log-odds variables $l_k$ as learnable parameters $\gamma_k \in \mathcal{O}$ initialized to $l_0$.
We can then re-write the map update equation  Eq.~\eqref{eq4:logodds_update} from OGM to match a parameter update equation from SGD as follows:
\begin{equation}
    \gamma_k := \gamma_k - \alpha \nabla \mathcal{L}(\mathcal{O}, z_n)
\end{equation}
where $\alpha$ is the learning rate and $\mathcal{L}$ is an objective function operating on the parameters $\mathcal{O}$ and a measurement $z_n$.
Matching the two equations, we get:
\begin{equation}
    -\alpha \nabla \mathcal{L}(\mathcal{O}, z_n) = \mathcal{S}(k, z_n) - \gamma_0
\end{equation}
Under this new setting, performing OGM translates to optimizing an unknown objective function $\mathcal{L}$, whose gradients play the role of the desired map update provided by the inverse measurement model $\mathcal{S}$. 

A trained OGM, $\mathcal{O}$, can be used to efficiently sample points along sensor rays $\vec{r}(t)$ during volumetric rendering in the following manner. 
Given the ray origin $\vec{o}$ and direction $\vec{d}$, we uniformly sample $N/2$ points $\vec{\zeta}_j = \vec{o} + \zeta_j \vec{d}$ along this ray where $N$ is the total number of samples needed for volumetric rendering. 
We compute the log-odds values at these continuous coordinates, $l_{\vec{\zeta}_j}$, using trilinear interpolation of the discrete grid $\mathcal{O}$. 
The corresponding occupancy is then computed as 
\begin{equation}
    p_{\vec{\zeta}_j} = \frac{1}{1 + \exp(-l_{\vec{\zeta}_j})}
\end{equation}
Note that a value of $p=0.5$ corresponds to unknown regions.
We clamp and re-scale these probabilities from $[0.5, 1.0]$ to $[0.0, 1.0]$ and perform importance sampling to get $N/2$ additional samples around occupied regions. 
Finally, both the uniform samples and the additional samples are concatenated to construct $N$ samples $\vec{\beta}_j = \vec{o} + \beta_j \vec{d}$ that are then used in volumetric rendering. 

In order to train $\mathcal{O}$, we define a function $g$ that defines the gradients of $l_{\vec{\beta}_j}$ as follows:
\begin{equation}
\begin{split}
    \frac{\partial \mathcal{L}}{\partial l_{\vec{\beta}_j}} =  g(\beta_j, z_n) &= l_{\mathtt{free}} \mathcal{U}((z_t - \delta) - \beta_j) + \\
                    & - l_{\mathtt{occ}} \mathcal{U}(\beta_j - (z_t - \delta)) \mathcal{U}((z_t + \delta) - \beta_j)
\end{split}
\end{equation}
where $\mathcal{U}$ is the Heaviside step function, which takes the value of 1 for positive inputs and zero otherwise, and $\delta$ is a hyperparameter.  
These gradients can then be backpropagated through the trilinear interpolation operator to compute updates for $\gamma_k$, as shown pictorially in Panel B of Fig.~\ref{fig:methods_occ_grid}, which shows an illustration of $g$.
Note that we do not explicitly define $\mathcal{L}(\mathcal{O})$.

\subsection{Training Details}
We perform training of the two NeRF MLPs in three stages. In Stage 1, we only train the sigma MLP using LiDAR data for $2500$ iterations with a batch size of $1024$ rays per iteration. Note that rendering a LiDAR ray is significantly faster than rendering a camera ray since only the sigma MLP needs to be evaluated. In Stage 2, we freeze the weights of the sigma MLP and train the color MLP using camera data for $2500$ iterations with $1024$ rays per iteration. Finally, in Stage 3, we jointly fine-tune the full model for $10000$ iterations with $1024$ LiDAR rays and $1024$ camera rays per iteration. In stages 1 and 3, we also train the OGM using the same batch of LiDAR data. We make one update with the OGM optimizer after accumulating gradients every 10 updates to the NeRF model. With the above settings, the model takes less than 12 minutes and 45 seconds 
to train on a single NVIDIA A100 GPU. 
%%  EXPERIMENTS - NOVEL VIEW SYNTHESIS FIGURE
\begin{figure*}[th]
   \centering
    \includegraphics[width=\linewidth]{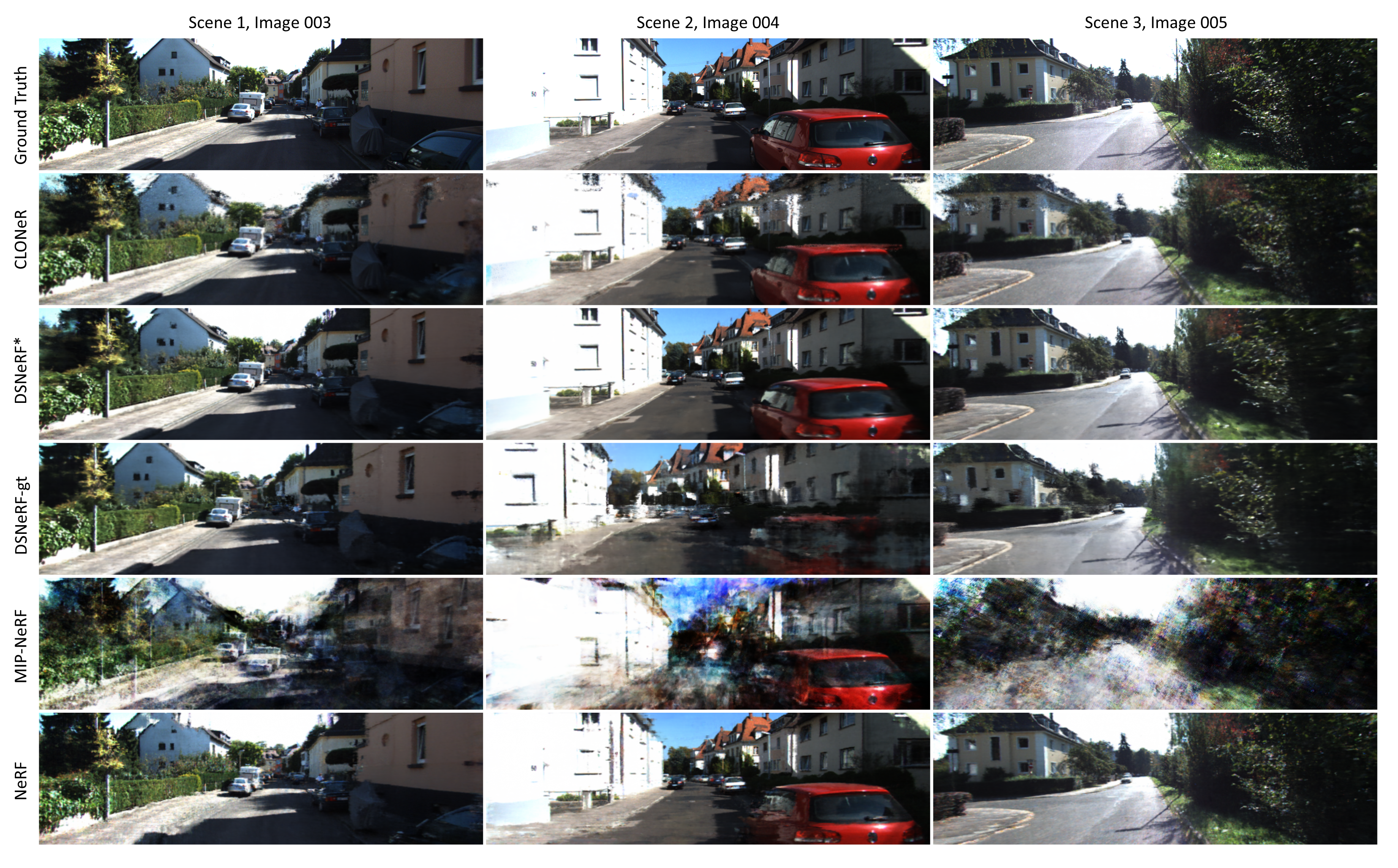}
    \caption{Qualitative comparison of novel view synthesis results for 3 different scenes. Best viewed using a monitor, please zoom in to view details.}
    \label{fig:qualitative_recon_results}
\end{figure*}

%% 3D RECON IMAGE FIGURE
\begin{figure*}[th]
   \centering
   \includegraphics[width=0.97\linewidth]{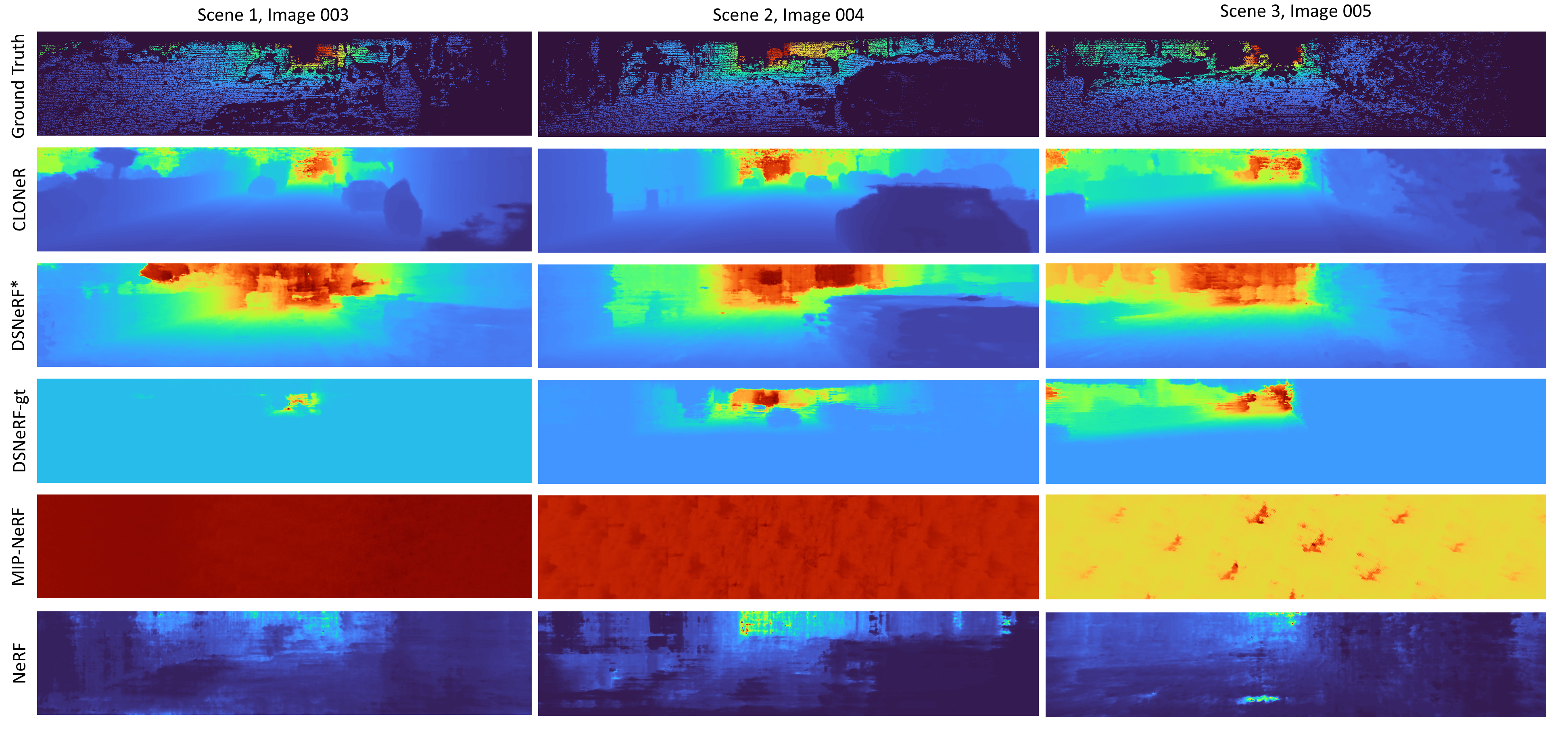}
    \caption{Qualitative comparison of rendered depth for the scenes corresponding to Fig.~\ref{fig:qualitative_recon_results}. Ground truth depth is from the KITTI depth prediction benchmark. The top 1/3rd of the depth maps, which is not covered by LiDAR, have been cropped out, as done in the KITTI depth estimation benchmark. Best viewed using a monitor, please zoom in to view details.}
    \label{fig:qualitative_3Drecon_results}
\end{figure*}

\section{Experimental Results}
\label{sec:results}
%% EXPERIMENTS - NOVEL VIEW SYNTHSIS QUANT RESULTS TABLE
\begin{table}[t]
    \centering
    \caption{Quantitative results for novel view synthesis. Metrics reported are averages across the 19 scenes.}
    \begin{tabular}{|c|c|c|c|}
    \hline
    Algorithm  & PSNR $\uparrow$ & MSSIM $\uparrow$ & LPIPS $\downarrow$  \\
    \hline
    NeRF &  $3.123$ & $0.169$ & $0.788$ \\
    MIP-NeRF & $13.274$ & $0.500$ & $0.255$ \\
    DSNeRF-gt & $12.849$ & $0.511$ & $0.268$  \\
    DSNeRF* & $19.932$ & $0.856$ & \textbf{0.120}  \\    
    CLONeR & \textbf{20.042} & \textbf{0.856} & 0.128  \\
    \hline
    \end{tabular}
    \label{tab4:quantitative_results_scenerecon_rgb}
\end{table}

%% 3D RECON QUANT RESULTS TABLE
\begin{table}[t]
    \centering
    \caption{Quantitative results for dense depth prediction. Metrics reported are averages across the 19 scenes. Ground truth depth is from the KITTI depth prediction benchmark. Regions of the scene that do not include LiDAR are not considered, as is done in the KITTI depth estimation benchmark.}
    \begin{tabular}{|c|c|c|c|c|c|c|}
    \hline
     Algorithm  & $\mu$SILog $\downarrow$ & absErrRel $\downarrow$ & sqErrRel $\downarrow$  \\
     \hline
    NeRF &  $0.417$ & $0.308$ & $0.568$ \\
    MIP-NeRF & $0.568$ & $0.405$ & $0.266$ \\
    DSNeRF-gt & $0.442$ & $0.365$ & $0.232$ \\
    DSNeRF* & $0.196$ & $0.085$ & $0.027$ \\    
    CLONeR & \textbf{0.101} & \textbf{0.073} & \textbf{0.017}  \\
    \hline
    \end{tabular}
    \label{tab4:quantitative_results_depthscenerecon_3D}
\end{table} 

In this section we present image-only results from for  Novel View Synthesis and Depth Prediction tasks. For video results of Fig.~\ref{fig:qualitative_recon_results} and Fig.~\ref{fig:qualitative_3Drecon_results}, please refer to the project website.

\subsection{Dataset and Training/Evaluation Protocol}
\label{subsec:dataset}

We train and evaluate the proposed model and baselines on 19 scenes from the KITTI dataset~\cite{geiger2012we}. Each scene consists of three rectified stereo image pairs and three LiDAR scans corresponding to consecutive time steps, $t-1, t$ and $t+1$. For training each baseline, we use camera poses and intrinsics estimated by COLMAP~\cite{schoenberger2016mvs,schoenberger2016sfm}. Since our proposed method operates in metric space, we use the pose and calibration information provided by the KITTI dataset to train CLONeR. For each scene, the two LiDAR scans and two left camera images at time $t-1$ and $t+1$, referred to as Image 000 and Image 002, are used for training. 
Note that only CLONeR uses the two LiDAR scans corresponding to time $t-1$ and $t+1$ for training. 
Quantitative and qualitative evaluation is performed on the remaining four images - left image at time $t$ and all three right images. We refer to these test images as Image 001, Image 003, Image 004, and Image 005, respectively. 

\subsection{Baseline methods}
 We compare our method against the following state-of-the-art methods: NeRF~\cite{mildenhall2020nerf}, Mip-NeRF~\cite{barron2021mip}, and DSNeRF~\cite{deng2022depth}. We train each of these models using the provided hyperparameters within each of the cited code bases. For NeRF, we used the original Tensorflow implementation. For Mip-NeRF, we use the available pytorch-lightning implementation ~\cite{mipnerfgithub}. For DSNeRF, we used the original pytorch implementation~\cite{deng2022depth}. Note that DSNeRF is trained using 3D keypoints estimated by COLMAP computed on all 6 images. In this setting, COLMAP would have likely used information from the 4 test images as well to estimate 3D keypoints visible from the 2 training images. We attempted to train DSNeRF using only keypoints that were estimated from the 2 training images but the training did not converge. This gives an advantage to DSNeRF compared to the other methods. We add an asterisk to all DSNeRF results to denote this advantage.
 In addition, we consider a fourth baseline, which we call DSNeRF-gt, where we replace pose, focal length and 3D keypoint information from COLMAP with ground truth information provided by the KITTI dataset. 
 Specifically, we aggregate points from the two LiDAR scans and treat them as 3D keypoints. We projected the points onto each frame and retained only points within its field of view. Note that DSNeRF-gt and our proposed method use the same amount of ground truth input information.

\vspace{-8pt}
\subsection{Novel View Synthesis Results}
We report three standard metrics for quantitative evaluation in Table~\ref{tab4:quantitative_results_scenerecon_rgb}: Peak Signal to Noise Ratio (PSNR)~\cite{korhonen2012peak}, Multiscale Structural Similarity Index (MSSIM)~\cite{wang2003multiscale}, and LPIPS~\cite{zhang2018perceptual}.
CLONeR meets or exceeds DSNeRF* with respect to PSNR and MSSIM while scoring marginally lower with respect to LPIPS. 
Note that CLONeR expects LiDAR data by design and is therefore at a disadvantage for the upper regions of the image where there is no LiDAR coverage. 
Qualitative results on selected images from three scenes are shown in Fig.~\ref{fig:qualitative_recon_results}.
While all methods were able to recover general scene details, Mip-NeRF performed significantly worse than others. Both CLONeR and DSNeRF* achieve similar level of detail in the rendered novel views. 
In our experiments, DSNeRF-gt performed worse than DSNeRF*. Although LiDAR-based depth is more precise than values obtained from COLMAP, different points in the LiDAR would be occluded from each of the camera viewpoint. And when projected, it associates a wrong depth to a scene point. As a result, we see that the trained model learns the image at $t-1$ and $t+1$ to be on two different planes.

\subsection{Depth Prediction Results}
For quantitative evaluation, we use the evaluation protocol from the KITTI depth prediction benchmark. This means that both our reported metrics and ground truth depth maps are from the KITTI depth prediction benchmark suite, and regions of the scene that do not include LiDAR are not considered, such as the sky.
While CLONeR and DSNeRF-gt operate in metric space, all the other baselines can only estimate depth up-to scale (as determined by the COLMAP poses). Thus, we follow the evaluation protocol in~\cite{wang2021can} and align the median depth of the predicted depth maps with the ground truth before computing the metrics. As seen in Table~\ref{tab4:quantitative_results_depthscenerecon_3D}, CLONeR outperforms all baselines across the three metrics when averaged across the 19 scenes.

The pseudo-ground truth depth maps used in the quantitative evaluation along with qualitative results are shown in
Fig.~\ref{fig:qualitative_3Drecon_results}.
Note that the pseudo-ground truth depth maps are sparse and contain many empty pixels where there is no ground truth. 
A visual inspection of the qualitative results demonstrate the improved performance of CLONeR compared to all the baselines. CLONeR captures greater details in all scenes along with fine structures in the regions where LiDAR data is present.
MIP-NeRF suffers from significant planar artifacts, potentially from the geometry ambiguity failure mode that NeRF models experience when trained on sparse views~\cite{barron2021mip360}. 
The depth maps from NeRF lack any structural detail.
For DSNeRF-gt, the left training image (Image 000) seems to be learnt on a plane close to the camera. As a result, the depth map for Image 003 (corresponding to right test image) has large planar portions that is behind the camera pose for the second training image (Image 002). Consequently, depth maps for Image 004 and Image 005 show more structural detail.
While DSNeRF* is able to capture global structure in the depth maps, the object boundaries are not clearly defined, see the results for scene 2, column 2 in the figure. It also struggles in regions with fine structures, such as the bushes in scene 3, final column of Fig.~\ref{fig:qualitative_3Drecon_results}.

\subsection{Ablation Experiments}

\begin{table}[!th]
    \centering
    \caption{Ablation experiments. Decoupled indicates whether the proposed decoupled Color/Sigma MLPs are used (instead of a single MLP) for fine sampling of the scene, LiDAR indicates if LiDAR rays are used as depth guidance, LOS/Term indicates if Line of Sight loss or Termination depth loss is used, and CF indicates if a coarse MLP is used (instead of the OGM).}
    \begin{tabular}{|c|c|c|c|c|c|}
    \hline
    Type & Decoupled & LiDAR & LOS / Term & OGM & CF  \\
    \hline
    Ablation A & \xmark & \xmark & N/A & \xmark & \cmark \\
    \hline
    Ablation B & \xmark & \cmark & \cmark / \xmark & \cmark & \xmark \\
    \hline
    Ablation C & \cmark & \cmark & \xmark / \cmark & \cmark & \xmark    \\
    \hline
    Ablation D & \cmark & \cmark & \cmark / \xmark & \xmark & \cmark \\
    \hline
    Proposed & \cmark & \cmark & \cmark / \xmark & \cmark & \xmark \\
    \hline
    \end{tabular}
    \label{tab1:ablation_configs}
\end{table}

%% FAILURE MODES FIGURE
\begin{figure}[t]
   \centering
   \includegraphics[width=0.97\linewidth]{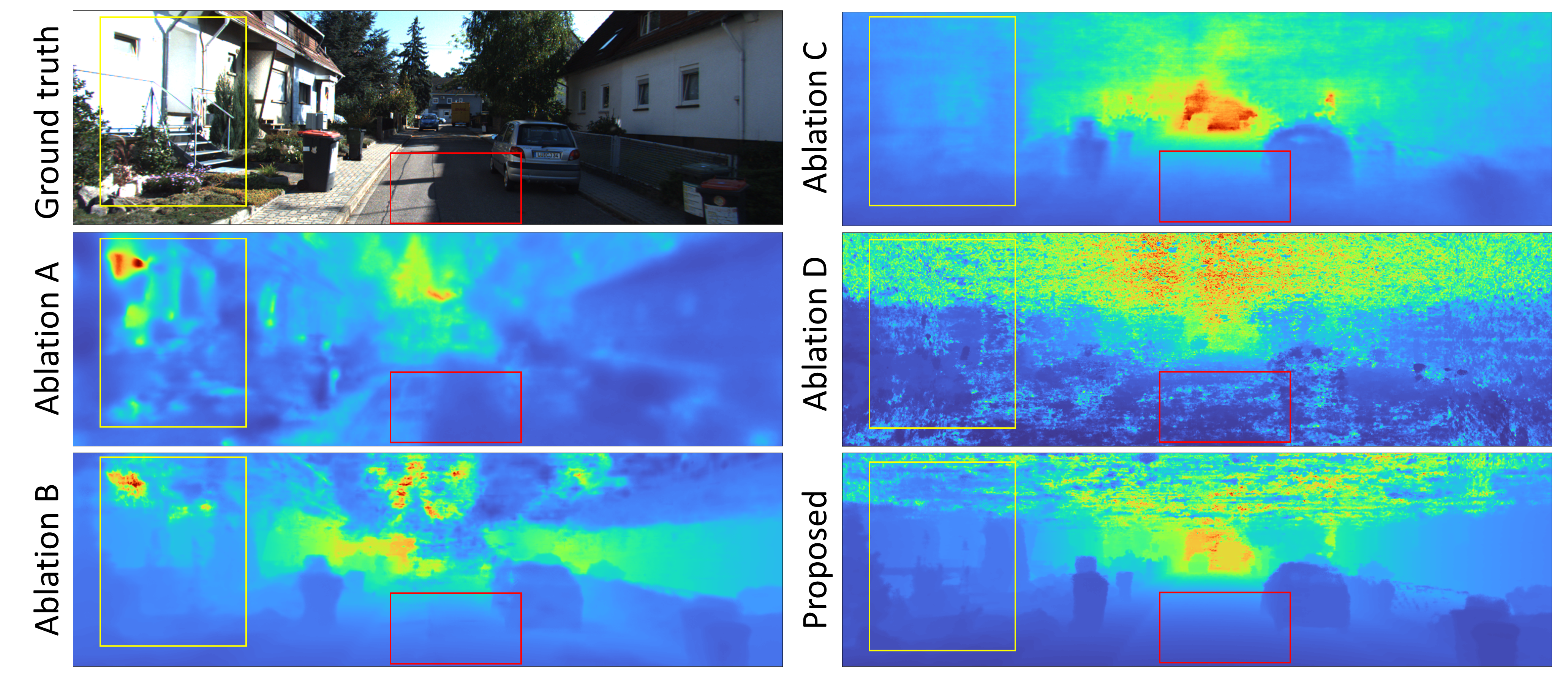}
    \caption{Qualitative comparison of rendered depth maps from a test image for different ablations. The yellow boxes highlight regions of saturation and red boxes highlight hard cast shadows}
    \label{fig:failuremodes_ablations}
\end{figure}

\begin{table}[th!]
    \centering
    \caption{Ablation quantitative evaluation for Novel View Synthesis and Depth estimation. Metrics reported are averages across the 19 scenes. Regions of the scene that do not include LiDAR are not considered, as is done in the KITTI depth estimation benchmark.}
    \begin{tabular}{|c|c|c|c|}
    \hline
    \multirow{2}{4em}{Ablation} & \multicolumn{3}{c|}{Novel View Synthesis} \\
    \cline{2-4}
        & PSNR $\uparrow$ & MSSIM $\uparrow$ & LPIPS $\downarrow$  \\
    \hline
        Ablation A & $18.491$ & $0.781$ & $0.148$ \\
        Ablation B & $15.803$ & $0.742$ & $0.179$ \\
        Ablation C & $17.142$ & $0.735$ & $0.189$ \\
        Ablation D & $17.657$ & $0.782$ & $0.156$  \\
        Proposed & ~\textbf{20.042} & ~\textbf{0.856} & ~\textbf{0.128}  \\
    \hline
    \hline
    \multirow{2}{4em}{Ablation} & \multicolumn{3}{c|}{Depth Prediction} \\
    \cline{2-4}
        &  SILog $\downarrow$ & absErrRel $\downarrow$ & sqrErrRel $\downarrow$ \\
    \hline
        Ablation A  & $0.554$ & $0.461$ & $0.422$ \\
        Ablation B  & $0.282$ & $0.141$ & $0.048$ \\
        Ablation C  & $0.132$ & $0.097$ & $0.027$ \\
        Ablation D & $0.353$ & $0.314$ & $0.264$ \\
        Proposed & \textbf{0.101} & \textbf{0.073} & \textbf{0.017} \\
    \hline
    \end{tabular}
    \label{tab:quantitative_results_ablation}
\end{table}

To analyze the different components of the proposed work, we consider 5 ablation experiments, described in Table~\ref{tab1:ablation_configs}. 
We present the quantitative results in Table~\ref{tab:quantitative_results_ablation}. The proposed method demonstrates comparable performance across ablation experiments for novel view synthesis, but yields the highest performance for depth prediction. 
In Fig.~\ref{fig:failuremodes_ablations}, we show how the different features of CLONeR eliminate the failure modes that NeRFs have in outdoor scenes.
Ablation A and B, which both use coupled NeRFs, experience the failure mode where both saturation and hard cast shadows are learned as separate objects or planes; these regions are highlighted in the yellow callout box and red callout box (respectively) in each depth map of Fig.~\ref{fig:failuremodes_ablations}.
Ablation A, B, and C all experience the failure mode where fine structures cannot be learned. We observe that all the structures in the yellow callout box for these depth maps are noticeably blurry, which highlights not only the importance of LiDAR but also the proposed volumetric carving losses. Ablation D experiences the failure mode where both global and local smooth scene structure is difficult to learn without proper sampling in a large outdoor scene, highlighting the importance of the proposed OGM. Finally, we observe that all ablation models struggle to learn correct scene geometry in the upper parts of the image.

\section{Conclusion}
\label{sec:conclusion}

This paper tackles the problem of novel view synthesis and dense depth prediction for large unbounded outdoor scenes with sparse input views. We proposed CLONeR, which extends NeRFs to operate on camera and LiDAR data and leverages occupancy grid mapping for efficient sampling, thus replacing the standard coarse NeRF MLP. Through both quantitative and qualitative experiments on the KITTI dataset, we demonstrate that the proposed method outperforms state-of-the-art algorithms on both tasks.
Future work will investigate improving the performance of the proposed method in regions where LiDAR data is not present, but camera data is available, namely the sky and regions of the scene where LiDAR cannot be collected, such as through glass. Ultimately, this method opens up significant opportunity to begin using NeRFs to learn dense maps of outdoor scenes and can allow researchers to augment existing datasets with novel views and corresponding depth maps. 

%\clearpage

% The acknowledgments are automatically included only in the final and preprint versions of the paper.
% \acknowledgments{If a paper is accepted, the final camera-ready version will (and probably should) include acknowledgments. All acknowledgments go at the end of the paper, including thanks to reviewers who gave useful comments, to colleagues who contributed to the ideas, and to funding agencies and corporate sponsors that provided financial support.}

\bibliographystyle{IEEEtran}
\bibliography{mainbib}

%\printbibliography
\end{document}